\documentclass[runningheads]{llncs}
\usepackage{graphicx}
\usepackage{tikz}
\usepackage{amsmath,amssymb} % define this before the line numbering.
\usepackage{comment}
\usepackage{color}

\usepackage{adjustbox}
\usepackage{booktabs}
\usepackage{multirow}
\usepackage{capt-of}
\usepackage{pifont}
\usepackage{breakcites}
\usepackage{subfiles}

\usepackage{algpseudocode}
\usepackage{algorithmicx,algorithm}

\usepackage{xcolor}
\usepackage{hyperref}
\hypersetup{
	colorlinks,
	linkcolor={red!60!black},
	citecolor={blue!30!black},
	urlcolor={blue!80!black}
}

\newcommand{\etal}{\textit{et al}.}
\newcommand{\ie}{\textit{i}.\textit{e}.}
\newcommand{\eg}{\textit{e}.\textit{g}.}
\newcommand{\etc}{\textit{etc}.}
\newcommand{\mysim}{\mathrm{sim}}

\begin{document}
% \linenumbers
\pagestyle{headings}
\mainmatter
\def\ECCVSubNumber{100}  % Insert your submission number here

\title{Self-supervised Video Representation Learning by Pace Prediction} % Replace with your title

\begin{comment}
\titlerunning{ECCV-20 submission ID \ECCVSubNumber} 
\authorrunning{ECCV-20 submission ID \ECCVSubNumber} 
\author{Anonymous ECCV submission}
\institute{Paper ID \ECCVSubNumber}
\end{comment}
%******************

%\begin{comment}
\titlerunning{Self-supervised Video Representation Learning by Pace Prediction} 
\author{Jiangliu Wang\inst{1} \and
Jianbo Jiao\inst{2} 
\and
{Yun-Hui Liu\inst{1}}\thanks{Corresponding author.}}
\authorrunning{J. Wang et al.}

\institute{The Chinese University of Hong Kong
\and
University of Oxford\\
\email{\{jlwang,yhliu\}@mae.cuhk.edu.hk~}, \email{jianbo@robots.ox.ac.uk}}

%\end{comment}
%******************
\maketitle

\begin{abstract}
This paper addresses the problem of self-supervised video representation learning from a new perspective -- by video pace prediction. It stems from the observation that human visual system is sensitive to video pace, \eg, slow motion, a widely used technique in film making. Specifically, given a video played in natural pace, we randomly sample training clips in different paces and ask a neural network to identify the pace for each video clip. The assumption here is that the network can only succeed in such a pace reasoning task when it understands the underlying video content and learns representative spatio-temporal features. In addition, we further introduce contrastive learning to push the model towards discriminating different paces by maximizing the agreement on similar video content. To validate the effectiveness of the proposed method, we conduct extensive experiments on action recognition and video retrieval tasks with several alternative network architectures. Experimental evaluations show that our approach achieves state-of-the-art performance for self-supervised video representation learning across different network architectures and different benchmarks. 
The code and pre-trained models are available at \url{https://github.com/laura-wang/video-pace}.
\keywords{Self-supervised learning \and Video representation \and Pace.}
\end{abstract}

\section{Introduction}

Convolutional neural networks have witnessed absolute success in video representation learning~\cite{carreira2017quo,tran2018closer,feichtenhofer2019slowfast} with human-annotated labels. Researchers have developed a wide range of neural networks~\cite{simonyan2014two,tran2015learning,tran2018closer} ingeniously, which extract powerful spatio-temporal representations for video understanding. Meanwhile, millions of labeled training data~\cite{karpathy2014large,kay2017kinetics} and powerful training resources are also the fundamental recipes for such great success. However, obtaining a large number of labeled video samples requires massive human annotations, which is expensive and time-consuming. Whereas at the same time, billions of unlabeled videos are available freely on the Internet. Therefore, video representation learning from unlabeled data is crucial for video understanding and analysis.

Among all the unsupervised learning paradigms, self-supervised learning is proved to be one promising methodology~\cite{oord2018representation,alwassel2019self}. The typical solution is to propose appropriate \emph{pretext tasks} that generate free training labels automatically and encourage neural networks to learn transferable semantic spatio-temporal features for the \textit{downstream tasks}. Such pretext tasks can be roughly divided into two categories: (1) generative dense prediction, such as flow fields prediction~\cite{gan2018geometry}, future frame prediction~\cite{srivastava2015unsupervised,vondrick2016generating}, \etc~ (2) discriminative classification/regression, such as video order prediction~\cite{misra2016shuffle,fernando2017self,lee2017unsupervised,xu2019self,kim2018self}, rotation transformation prediction~\cite{jing2018self}, motion and appearance statistics regression \cite{wang2019self}, \etc~
While promising results have been achieved, some of the approaches leverage pre-computed motion channels~\cite{wang2019self,gan2018geometry}, \eg, optical flow, to generate the training labels. This could be both time and space consuming, especially when the pre-training dataset scales to millions/trillions of data. To alleviate such a problem, in this work, we propose a simple yet effective pretext task without referring to pre-computed motion. Instead, we only base on the original videos as input.

\begin{figure}[t]
	\includegraphics[width=1\textwidth]{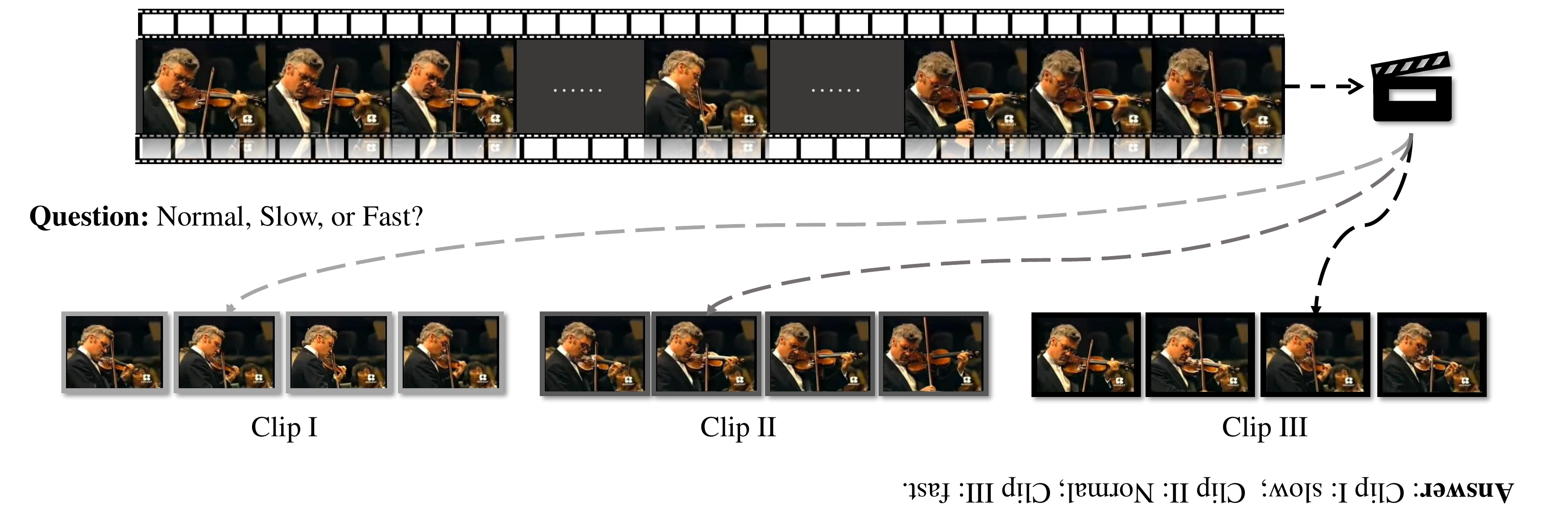}
	\caption{Illustration of the proposed pace prediction task. Given a video sample, frames are randomly selected by different paces to formulate the training inputs. Here, three different clips, \emph{Clip I, II, III}, are sampled by normal, slow and fast pace randomly. Can you ascribe the corresponding pace label to each clip? (Answer is provided below.)}
	\label{fig:tease}
\end{figure}

Inspired by the rhythmic montage in film making, we observe that human visual system is sensitive to motion pace and can easily distinguish different paces once understanding the covered content. Such a property has also been revealed in neuroscience studies~\cite{giese2003neural,watamaniuk1992human}.
To this end, we propose a simple yet effective task to perform self-supervised video representation learning: pace prediction. 
Specifically, given videos played in natural pace, video clips with different paces are generated according to different temporal sampling rates.
A learnable model is then trained to identify which pace the input video clip corresponds to. As aforementioned, the assumption here is that if the model is able to distinguish different paces, it has to understand the underlying content.
Fig. \ref{fig:tease} illustrates the basic idea of the proposed approach.

In the proposed pace prediction framework, we utilize 3D convolutional neural networks (CNNs) as our backbone network to learn video representation, following prior works~\cite{xu2019self,luo2020video}. Specifically, we investigated several architectures, including C3D~\cite{tran2015learning}, 3D-ResNet~\cite{tran2018closer,hara2018can}, R(2+1)D~\cite{tran2018closer}, and S3D-G~\cite{xie2018rethinking}.
Furthermore, we incorporate contrastive learning to enhance the discriminative capability of the model for video understanding.
Extensive experimental evaluations with several video understanding tasks demonstrate the effectiveness of the proposed approach. We also present a study of different backbone architectures as well as alternative configurations of contrastive learning. The experimental result suggests that the proposed approach can be well integrated into different architectures and achieves state-of-the-art performance for self-supervised video representation learning.

The main contributions of this work are summarized as follows.
\begin{itemize}
	\item We propose a simple yet effective approach for self-supervised video representation learning by pace prediction. This novel pretext task provides a solution to learn spatio-temporal features without explicitly leveraging the motion channel, \eg, optical flow.              
	\item We further introduce contrastive learning to regularize the pace prediction objective. Two configurations are investigated by maximizing the mutual information either between same video pace or same video context.    
	\item Extensive experimental evaluations on three network architectures and two downstream tasks across three datasets show that the proposed approach achieves state-of-the-art performance and demonstrates great potential to learn from tremendous amount of video data, in a simple manner. 
\end{itemize}

% Note that for clarity, we only use four frames  

\section{Related Work}
% We briefly review the most related work, including video representation learning and self-supervised learning. 
%Readers may refer to \cite{kolesnikov2019revisiting,jing2019self} for a more detailed review on self-supervised visual representation learning.
 
\subsubsection{Video Representation Learning.}
Video understanding, especially action recognition, has been extensively studied for decades, where video representation learning serves as the fundamental problem of other video-related tasks, such as complex action recognition~\cite{hussein2019timeception}, action temporal localization~\cite{chao2018rethinking,shou2017cdc,shou2016temporal}, video caption~\cite{wang2018reconstruction,wang2018bidirectional}, \etc~  
Initially, various hand-crafted local spatio-temporal descriptors are proposed for video representations, such as STIP \cite{laptev2005space}, HOG3D \cite{klaser2008spatio}, \etc~Wang \etal~\cite{wang2013action} proposed improved dense trajectories (iDT) descriptors, which combined the effective HOG, HOF~\cite{laptev2008learning} and MBH descriptors~\cite{dalal2006human}, and achieved the best results among all hand-crafted features. With the impressive success of CNN in image understanding problem and the availability of large-scale video datasets such as sports1M~\cite{karpathy2014large}, ActivityNet~\cite{caba2015activitynet}, Kinetics-400~\cite{carreira2017quo}, studies on data-driven deep learning-based video analysis started to emerge. According to the input modality, these video representation learning methods can be roughly divided into two categories: one is to directly take RGB videos as inputs, while the other is to take both RGB videos and optical flows as inputs. Tran \etal~\cite{tran2015learning} extended the 2D convolution kernels to 3D and proposed C3D network to learn spatio-temporal representations. Simonyan and Zisserman~\cite{simonyan2014two} proposed a two-stream network that extracts spatial features from RGB inputs and temporal features from optical flows, followed by a fusion scheme. Recently,~\cite{feichtenhofer2019slowfast} proposed to capture the video information in a slow-fast manner, which showed that input videos with slow and fast speed help the supervised action recognition task. 
Although impressive results have been achieved, video representation learning with human-annotated label is both time-consuming and expensive.

\subsubsection{Self-supervised Learning.} Self-supervised learning is becoming increasingly attractive due to its great potential to leverage the large amount of unlabeled data. The key idea behind it is to propose a pretext task that generates pseudo training labels without human annotation. Various pretext tasks have been proposed for self-supervised image representation learning, such as context-based prediction~\cite{doersch2015unsupervised}, rotation prediction~\cite{gidaris2018unsupervised}, colorization~\cite{zhang2016colorful}, inpainting~\cite{pathak2016context}, clustering~\cite{caron2018deep} and contrastive learning~\cite{oord2018representation,chen2020simple,henaff2019data,bachman2019learning}, to name a few.     

Recently, pretext tasks designed for videos are investigated to learn generic representations for downstream video tasks, such as action recognition, video retrieval, \etc~Intuitively, a large number of studies~\cite{fernando2017self,lee2017unsupervised,misra2016shuffle} leveraged the distinct temporal information of videos and proposed to use frame sequence ordering as their pretext tasks. B\"{u}chler \etal~\cite{buchler2018improving} further used deep reinforcement learning to design the sampling permutations policy for order prediction tasks. Gan \etal~\cite{gan2018geometry} proposed to learn video representations by predicting the optical flow or disparity maps between frames. Although these methods demonstrate promising results, the learned representations are only based on one or two frames as they used 2D CNN for self-supervised learning. Consequently, some recent works~\cite{han2019video,luo2020video,xu2019self,kim2018self} proposed to use 3D CNNs as backbone networks for spatio-temporal representations learning, among which~\cite{kim2018self,xu2019self,luo2020video} extended the 2D frame ordering pretext tasks to 3D video clip ordering, and \cite{han2019video} proposed a pretext task to predict future frames embedding. Some concurrent works~\cite{benaim2020speednet,yao2020video,jenni2020video} also investigate the speed property of video as in this work, and the reader is encouraged to review them for a broader picture. 
Self-supervised learning from multi-modality sources, \eg, video and audio~\cite{owens2018audio,korbar2018cooperative,alwassel2019self}, also demonstrated promising results.
%, which achieves comparable results with supervised learning on kinetics-400~\cite{kay2017kinetics}when using a large dataset IG65M~\cite{ghadiyaram2019large}. 
In this paper, we focus on the video modality only and leave the potential extension to multi-modality as future research.

\section{Our Approach}

% \subsection{Overview}
We address the video representation learning problem in a self-supervised manner. To achieve this goal, rather than training with human-annotated labels, we train a model with labels generated \textit{automatically} from the video inputs $X$. The essential problem is how to design an appropriate transformation $g(\cdot)$, usually termed as pretext task, so as to yield transformed video inputs $\widetilde{X}$ with human-annotated free labels that encourage the network to learn powerful semantic spatio-temporal features for the downstream tasks, \eg, action recognition. 
    
In this work, we propose pace transformation $g_{pac}(\cdot)$ with a pace prediction task for self-supervised learning. Our idea is inspired by the \textit{slow motion} which is widely used in film making for capturing a key moment and producing dramatic effect. Humans can easily identify it due to their sensitivity of the pace variation and a sense of normal pace. We explore whether a network could also have such ability to distinguish video play pace. Our assumption is that a network is not capable to perform such pace prediction task effectively unless it understands the video content and learns powerful spatio-temporal representations.

% In the following, we first elaborate on the pace prediction task. Then we introduce two possible contrastive learning strategies. Finally, we present the complete learning framework with three different 3D network architectures.           

\subsection{Pace Prediction}

\begin{figure}[t]
	\includegraphics[width=1\textwidth]{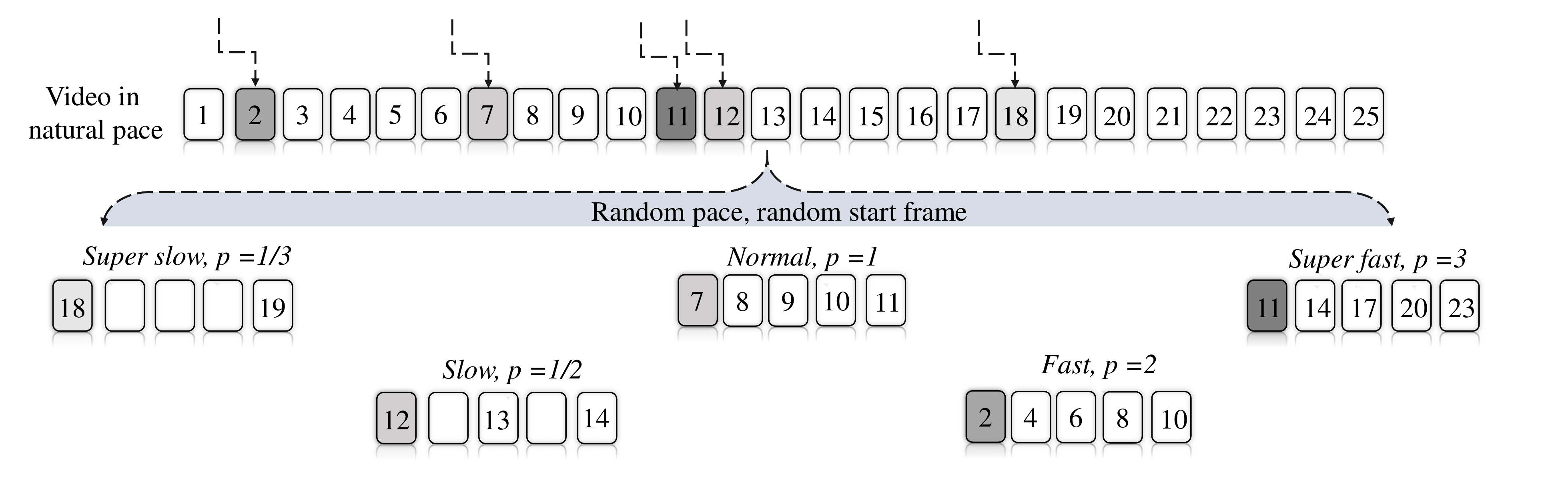}
	\caption{Generating training samples and pace labels from the proposed pretext task. Here, we show five different sampling paces, named as \emph{super slow}, \emph{slow}, \emph{normal}, \emph{fast}, and \emph{super fast}. The darker the initial frame is, the faster the entire clip plays.}
	\label{fig:sr}
\end{figure}

We aim to train a model with pace-varying video clips as inputs and ask the model to predict the video play paces. We assume that such a pace prediction task will encourage the neural network to learn generic transferable video representations and benefit downstream tasks.  Fig. \ref{fig:sr} shows an example of generating the training samples and pace labels. Note that in this example, we only illustrate one training video with five distinct sampling paces. Whereas in our final implementation, the sampling pace is randomly selected from several pace candidates, not restricted to these five specific sampling paces. 

As shown in Fig. \ref{fig:sr}, given a video in natural pace with 25 frames, training clips will be sampled by different paces $p$. Typically, we consider five pace candidates \{super slow, slow, normal, fast, super fast\}, where the corresponding paces $p$ are 1/3, 1/2, 1, 2, and 3, respectively. Start frame of each video clip is randomly generated and will loop over the video sample if the desired training clip is longer than the original video sample.
Methods to generate each training clip with a specific $p$ are illustrated in the following:
%We now go into each training clip to see how to generate them with a specific $p$ in the following: 

\begin{itemize}
	\item \emph{Normal motion}, where $p=1$, training clips are sampled consecutively from the original video. The video play speed is the same as the normal pace.
	
	\item \emph{Fast motion}, where $p>1$, we directly sample a video frame from the original video for every $p$ frames, \eg, super fast clip with $p=3$ contains frames 11, 14, 17, 20 and 23. As a result, when we play the clip in nature 25 fps, it looks like the video is speed up compared to the original pace. 
	
	\item \emph{Slow motion}, where $p<1$, we put the sampled frames into the five-frames clip for every $1/p$ frames instead, \eg, slow clip with $p=1/2$, only frames 1, 3, 5 are filled with sampled frames. Regarding the blank frames, one may consider to fill it with preceding frame, or apply interpolation algorithms \cite{jiang2018super} to estimate the intermediate frames. In practice, for simplicity, we use the preceding frame for the blank frames.
\end{itemize}

% While the interpolation of the blank frame is quite an interesting problem, in practice, we consider to use a \emph{relative} pace instead of using the \emph{absolute} pace, as the pace itself is also defined by the capturing system (\eg, camera models). For example, we assume clips with $p=3$ to be the anchor, \ie, normal pace and therefore, clips with $p>3$ are with faster pace and $p<3$ with slower pace. 

Formally, we denote the pace sampling transformation as $g(x)$.  Given a video $x$, we apply $g(x|p)$ to obtain the training clip $\widetilde{x}$ with a training pace $p$. The pace prediction pretext task is formulated as a classification problem and the neural network $f(\widetilde{x})$ is trained with cross entropy loss $\mathcal{L}_{cls}$ described as:
\begin{equation}
\mathcal{L}_{cls} = -\sum\limits_{i=1}^{M}y_i(\log\frac{\exp(h_i)}{\sum_{j=1}^{M}\exp(h_j)}), ~~~ h = f(\widetilde{x}) = f(g_{pac}(x|p)),\label{eq:loss_cls}
\end{equation}
% \begin{equation}
% \mathcal{L}_{cls} = -\sum\limits_{i=1}^{M}y_i(\log\frac{\exp(h_i)}{\sum_{j=1}^{M}\exp(h_j)}),\label{eq:loss_cls}
% \end{equation}
where M is the number of all the pace rate candidates.

\subsubsection{Avoid Shortcuts.} As first pointed out in \cite{doersch2015unsupervised}, when designing a pretext task, one must pay attention to the possibility that a network could be cheating or taking shortcuts to accomplish the pretext task by learning trivial solutions/low-level features rather than the desired high-level semantic representations.   
Such observations are also reported in~\cite{han2019video,kim2018self} for self-supervised video representation learning. 
In this work, we adopt color jittering augmentation to avoid such shortcuts. Empirically, we find that color jittering applied to each frame achieves much better performance than to the entire video clip. More details see Sec.~\ref{sec:ablation}.

\subsection{Contrastive Learning} 

To further enhance the pace prediction task and regularize the learning process, we propose to leverage contrastive learning as an additional objective.
Contrastive learning in a self-supervised manner has shown great potential and achieved comparable results with supervised visual representation learning recently \cite{oord2018representation,chen2020simple,bachman2019learning,henaff2019data,han2019video,wu2018unsupervised,tian2019contrastive,han2020memory}. It stems from Noise-Contrastive Estimation~\cite{gutmann2010noise} and aims to distinguish the positive samples from a group of negative samples. The fundamental problem of contrastive learning lies in the definition of \emph{positive} and \emph{negative} samples. For example, Chen~\etal~\cite{henaff2019data} consider the pair with different data augmentations applied to the same sample as positive, while Bachman~\etal~\cite{bachman2019learning} takes different views of a shared context as positive pair. 
In this work, we consider two possible strategies to define positive samples: same context and same pace. 
In the following, we elaborate on these two strategies. 

\subsubsection{Same Context.} We first consider to use clips from the same video but with different sampling paces as positive pairs, while those clips sampled from different videos as negative pair, \ie, content-aware contrastive learning.

Formally, given a mini-batch of $N$ video clips \{$x_1, \dots, x_N$\}, for each video input $x_i$, we randomly sample $n$ training clips from it by different paces, resulting in an actual training batch size $n*N$. Here, for simplicity, we consider $n=2$, and the corresponding positive pairs are \{($\widetilde{x}_i, p_i)$, (${\widetilde{x}_i}', {p_i}')$\}, where $\widetilde{x}_i$ and ${\widetilde{x}_i}'$ are sampled from the same video. Video clips sampled from different video are considered as negative pairs, denoted as \{($\widetilde{x}_i, p_i)$, (${\widetilde{x}_{\mathcal{J}}, p_\mathcal{J}}$)\}. Each video clip is then encoded into a feature vector $z_i$ in the latent space by the neural network $f(\cdot)$. Then the positive feature vector pair is $(z_i, {z_i}')$ while the negative pairs are \{($z_i,z_{\mathcal{J}})$\}. Denote $\mysim(z_i, {z_i}')$ as the similarity between feature vector $z_i$ and ${z_i}'$ and $\mysim(z_i, z_{\mathcal{J}}')$ as the similarity between feature vector $z_i$ and $z_{\mathcal{J}}'$, the content-aware contrastive loss is defined as:
\begin{equation}
\mathcal{L}_{ctr\_sc} = -\frac{1}{2N} \sum\limits_{i,\mathcal{J}} \log \frac{\exp (\mysim(z_i, {z_i}'))}{\sum\limits_{i} \exp(\mysim(z_i, {z_i}'))+ \sum\limits_{i,\mathcal{J}} \exp(\mysim(z_i, z_\mathcal{J}))},
\label{eq:ctr_sc}
\end{equation}
where $\mysim(z_i, {z_i}')$ is achieved by the dot product  ${z_i}^\top {z_i}'$ between the two feature vectors and so as $\mysim(z_i, z_{\mathcal{J}}')$.

\subsubsection{Same Pace.} Concerning the proposed pace prediction pretext task, another alternative contrastive learning strategy based on same pace is explored. Specifically, we consider video clips with the same pace as positive samples regardless of the underlying video content, \ie, content-agnostic contrastive learning. In this way, the contrastive learning is investigated from a different perspective that is explicitly related to pace.

Formally, given a mini-batch of $N$ video clips \{$x_1, \dots, x_N$\}, we first apply the pace sampling transformation $g_{pac}(\cdot)$ described above to each video input to obtain the training clips and their pace labels, denoted as \{($\widetilde{x}_1, p_1)$,\dots, ($\widetilde{x}_N, p_N)$\}. Each video clip is then encoded into a feature vector $z_i$ in the latent space by the neural network $f(\cdot)$. Consequently, $(z_i, z_j)$ is considered as positive pair if $p_i = p_j$ while $(z_i, z_k)$ is considered as negative pair if $p_i \neq p_k$, where $j, k \in \{1,2,\dots,N\}$. Denote $\mysim(z_i, z_j)$ as the similarity between feature vector $z_i$ and $z_j$ and $\mysim(z_i, z_k)$ as the similarity between feature vector $z_i$ and $z_k$, the contrastive loss is defined as:
\begin{equation}
\mathcal{L}_{ctr\_sp} = -\frac{1}{N} \sum\limits_{i,j,k} \log \frac{\exp (\mysim(z_i, z_j))}{\sum\limits_{i,j} \exp(\mysim(z_i, z_j))+ \sum\limits_{i,k} \exp(\mysim(z_i, z_k))},
\label{eq:ctr_sp}
\end{equation}
where $\mysim(z_i, z_j)$ is achieved by the dot product  ${z_i}^\top z_j$ between the two feature vectors and so as $\mysim(z_i, z_k)$.  
% While cosine similarity ${z_i}^\top z_j/||z_i||||z_j||$ may also be a possible solution, we empirically found that the dot product
%Recently, Chen \etal~also proposed to use a cosine similarity ${z_i}^\top z_j/||z_i||||z_j||$ to effectively weight different examples. While we don't consider to use it here as during the experiment, we find that the dot product can already give a good performance.

\subsection{Network Architecture and Training}
\begin{figure}[t]
	\includegraphics[width=1\textwidth]{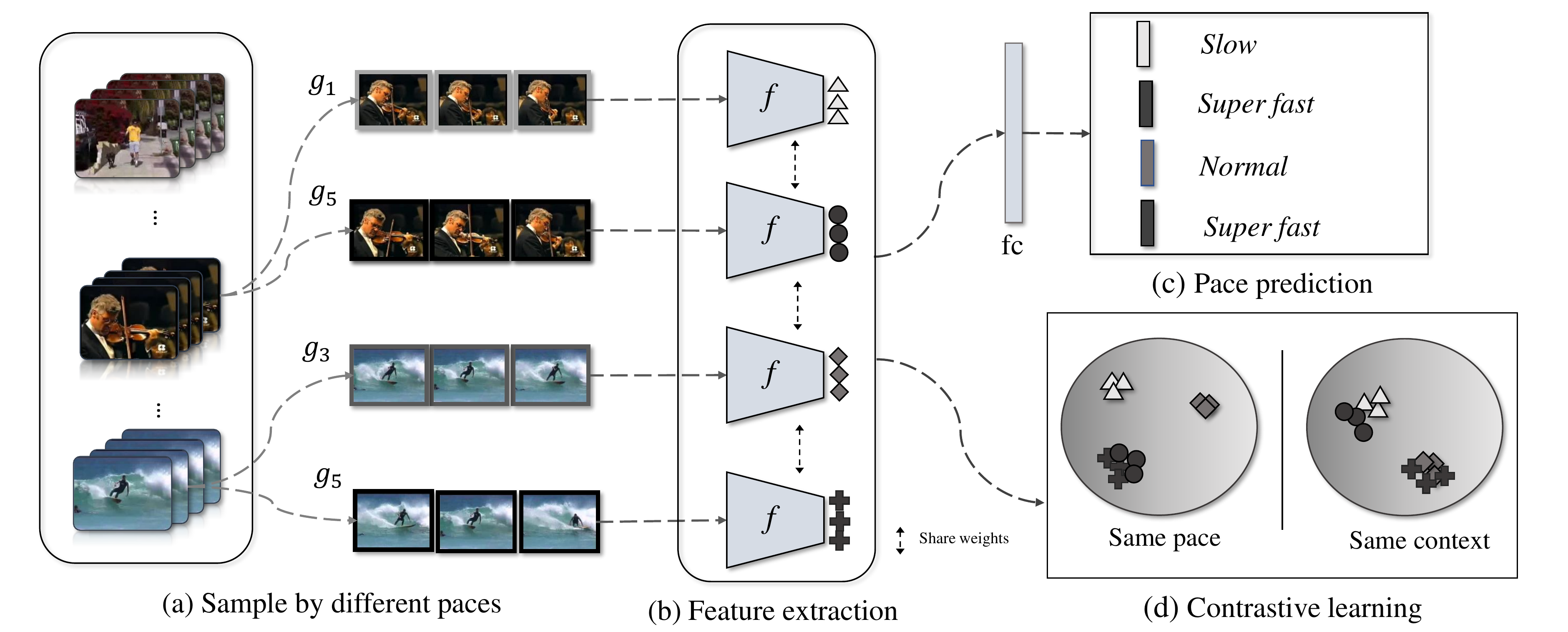}
	\caption{Framework of the proposed approach. (a) Training clips are sampled by different paces. Here, $g_1, g_3, g_5$ illustrates examples of \emph{slow}, \emph{normal} and \emph{super fast} pace. (b) A 3D CNN $f$ is leveraged to extract spatio-temporal features. (c) The model is trained to predict the specific pace applied to each video clip. (d) Two possible contrastive learning strategies are considered to regularize the learning process at the latent space. The symbols at the end of the CNNs represent feature vectors extracted from different video clips, where the intensity represents different video pace.}
	\label{fig:framework}
\end{figure}

The framework of the proposed pace prediction approach is illustrated in Fig. \ref{fig:framework}. Given a set of unlabeled videos, we firstly sample various video clips with different paces. Then these training clips are fed into a deep model (3D CNN here) for spatio-temporal representation learning.  The final objective is to optimize the model to predict the pace of each video clip and maximize the agreement (mutual information) between positive pairs at the same time.

In terms of the network architecture, we mainly consider three backbone networks, \ie~C3D~\cite{tran2015learning}, 3D-ResNet~\cite{hara2018can} and R(2+1)D~\cite{tran2018closer}, to study the effectiveness of the proposed approach. 
C3D is a classic neural network which operates on 3D video volume by extending 2D convolutional kernel to 3D. 3D-ResNet (R3D)~\cite{hara2018can} is an extension of the ResNet~\cite{he2016deep} architecture to videos. In this work, we use 3D-ResNet18 (R3D-18).
R(2+1)D~\cite{tran2018closer} is proposed to break the original spatio-temporal 3D convolution into a 2D spatial convolution and a 1D temporal convolution.
Apart from these three networks, we also use a state-of-the-art model S3D-G~\cite{xie2018rethinking} to further exploit the potential of the proposed approach.

By jointly optimizing the classification objective (Eq.~\ref{eq:loss_cls}) and the contrastive objective (Eq.~\ref{eq:ctr_sc} or \ref{eq:ctr_sp}), the final training loss is defined as:
\begin{equation}
\mathcal{L} = \lambda_{cls}\mathcal{L}_{cls} + \lambda_{ctr}\mathcal{L}_{ctr},
\end{equation}
where $\lambda_{cls}$, $\lambda_{ctr}$ are weighting parameters to balance the optimization of pace prediction and contrastive learning, respectively. The $\mathcal{L}_{ctr}$ refers to either contrastive learning with same pace $\mathcal{L}_{ctr\_sp}$ or with same context $\mathcal{L}_{ctr\_sc}$.

\section{Experiments}\label{sec.exp}

\subsection{Datasets and Implementation Details}

\subsubsection{Datasets.} We use three datasets as follows:

\emph{UCF101}~\cite{soomro2012ucf101} is a widely used dataset in action recognition task, consisting of 13,320 video samples with 101 action classes. 
%The covered actions are all naturally performed as they are collected from YouTube. 
The dataset is divided into three training/testing splits and in this paper, following prior works~\cite{xu2019self,benaim2020speednet}, we use training split 1 as pre-training dataset and training/testing split 1 for evaluation. 
 
\emph{Kinetics-400 (K-400)}~\cite{kay2017kinetics} is a large action recognition dataset, which consists of 400 human action classes and around 306k videos. It is divided into three splits: training/validation/testing. In this work, we use the training split (around 240k video samples) as the pre-training dataset, to validate the proposed approach.

\emph{HMDB51}~\cite{kuehne2011hmdb} is a relatively small action recognition dataset, which contains around 7,000 videos with 51 action classes. 
It is divided into three training/testing splits and following prior works~\cite{xu2019self,benaim2020speednet}, we use training/testing split 1 for downstream task evaluation. 

\subsubsection{Self-supervised Pre-training Stage.} When pre-training on the K-400 dataset, for each video input, we first generate a frame index randomly and then start from the index, sample a consecutive 16-frame video clip. While when pre-training on UCF101 dataset, as it only contains around 9k videos in the training split, we set epoch size to be around 90k for temporal jittering following~\cite{alwassel2019self}. As for data augmentation, we randomly crop the video clip to 112 $\times$ 112 and flip the whole video clip horizontally. The batch size is 30 and SGD is used as optimizer with an initial learning rate $1\times10^{-3}$. The leaning rate is divided by 10 for every 6 epochs and the training process is stopped after 18 epochs. When jointly optimizing $\mathcal{L}_{cls}$ and $\mathcal{L}_{ctr}$, $\lambda_{cls}$ is set to 1 and $\lambda_{ctr}$ is set to 0.1.   

\subsubsection{Supervised Fine-tuning Stage.} Regarding the action recognition task, during the fine-tuning stage, weights of convolutional layers are retained from the self-supervised learning networks while weights of fully-connected layers are randomly initialized. The whole network is then trained with cross-entropy loss. 
Image pre-processing method and training strategy are the same as the self-supervised pre-training stage, except that the initial learning rate is set to 0.01 for S3D-G and $3\times10^{-3}$ for the others.

\subsubsection{Evaluation.} During inference, following the evaluation protocol in~\cite{xu2019self,luo2020video}, we sample 10 clips uniformly from each video in the testing set of UCF101 and HMDB51. For each clip, center crop is applied. The predicted label of each video is generated by averaging the softmax probabilities of all clips in the video.  

\subsection{Ablation Studies}\label{sec:ablation}

\begin{table}[t]
	\begin{minipage}{.49\textwidth}
	    \caption{Explore the best setting for pace prediction task. Sampling pace $p=[a,b]$ represents that the lowest value of pace $p$ is $a$ and the highest is $b$ with an interval of 1, except $p=[\frac{1}{3},3]$, where $p$ is selected from $\{\frac{1}{3},\frac{1}{2}, 1, 2, 3\}$.}
	    
	    \label{tab:pace}
		\centering
		\begin{adjustbox}{max width=\textwidth}
		\begin{tabular}{cccc}
			\toprule
			Color jittering & Method & \#Classes & UCF101 \\
			\midrule
			$\times$  & Random & - &56.0 \\
			\midrule
			$\times$  & $p=[1,3]$ &3 & 71.4 \\
			$\times$  & $p=[1,4]$ &4 & \textbf{72.0} \\
			$\times$  & $p=[1,5]$ &5 &72.0 \\
			$\times$  & $p=[1,6]$ &6 & 71.1 \\
			\midrule
			\checkmark & $p=[1,4]$ & 4 &\textbf{73.9}\\
			\checkmark & $p=[\frac{1}{3},3]$ & 5 &73.9 \\
			\bottomrule
		\end{tabular}
		\end{adjustbox}
	\end{minipage}
	\hfill
	\begin{minipage}{.49\textwidth}
		\centering\includegraphics[width=0.9\linewidth]{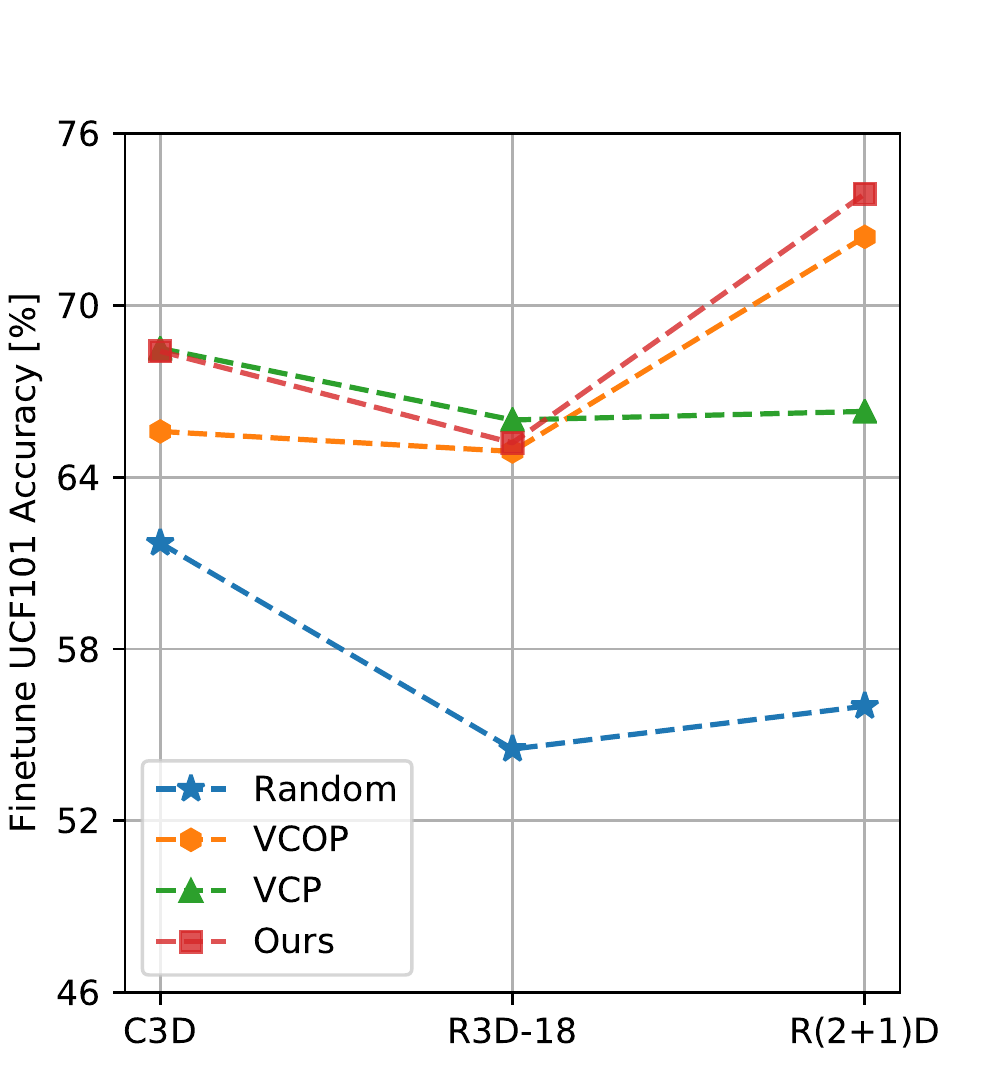}
		\captionof{figure}{Action recognition accuracy on three backbone architectures (horizontal axis) using four initialization methods.}
		\label{fig:backbone}
	\end{minipage}
\end{table}

In this section, we firstly explore the best sampling pace design for the pace prediction task. We apply it to three different backbone networks to study the effectiveness of the pretext task and the network architectures. Experimental results show that respectable performance can be achieved by only using the pace prediction task. When introducing the contrastive learning, the performance is further boosted, and the same content configuration performs much better than the same pace one.
More details are illustrated in the following.

\subsubsection{Sampling Pace.} We investigate the best setting for pace prediction task with R(2+1)D backbone network~\cite{tran2018closer} in Table \ref{tab:pace}. 
To study the relationship between the complexity of the pretext task and the effectiveness on the downstream task, we first use a \emph{relative} pace design with only normal and fast motion. For example, regarding $p=[1,3]$, we assume clips with $p=2$ to be the anchor, \ie, normal pace and therefore, clips with $p=3$ are with faster pace and $p=1$ are with slower pace.  
It can be seen from the table that with the increase of the maximum pace, namely the number of training classes, the accuracy of the downstream action recognition task keeps increase, until $p=[1,4]$. When the sampling pace increases to $p=[1,6]$, the accuracy starts to drop. We believe that this is because such a pace prediction task is becoming too difficult for the network to learn useful semantic features. This provides an insight on the pretext task design that a pretext task should not be too simple nor too ambiguous to solve, in consistent with the observations found in \cite{noroozi2016unsupervised,fernando2017self}. 

We further validate the effectiveness of color jittering based on the best sampling pace design $p=[1,4]$. It can be seen from Table~\ref{tab:pace} that with color jittering, the performance is further improved by 1.9\%.
It is also interesting to note that the \emph{relative} pace, \ie, $p=[1, 4]$, achieves comparable result with the \emph{absolute} pace, \ie, $p=[\frac{1}{3}, 3]$, but with less number of classes.
In the following experiments, we use sampling pace $p=[1,4]$ along with color jittering by default.

\subsubsection{Backbone Network.} We validate the proposed pace prediction task without contrastive learning using three backbone networks.  
Some recent works~\cite{xu2019self,luo2020video} validate their proposed self-supervised learning approaches on modern spatio-temporal representation learning networks, such as R3D-18~\cite{hara2018can,tran2018closer}, R(2+1)D~\cite{tran2018closer}, \etc~This practice could influence the direct evaluation of the pretext tasks, as the performance improvement can also come from the usage of more powerful networks. Therefore, we study the effectiveness of the pace prediction task and compare with some recent works on three backbone networks, as shown in Fig. \ref{fig:backbone}. For a fair comparison, following~\cite{xu2019self,luo2020video}, we use the first training split of UCF101 as the pre-training dataset and evaluate on training/testing split 1.      

Some key observations are listed in the following: (1) The proposed approach achieves significant improvement over the random initialization across all three backbone networks.
(2) Although in the random initialization setting, C3D achieves the best results, R(2+1)D and R3D-18 benefit more from the self-supervised pre-training and R(2+1)D finally achieves the best performance. (3) Without contrastive learning, the proposed pace prediction task already demonstrates impressive effectiveness to learn video representations, achieving comparable performance with current state-of-the-art methods VCP~\cite{luo2020video} and VCOP\cite{xu2019self} on C3D and R3D-18 and outperforms them when using R(2+1)D. 

\begin{table}[tbp!]
    \centering
	\caption{Evaluation of different contrastive learning configurations on both UCF101 and HMDB51 datasets. $^*$Note that paramters when adding a fc layer only increase $\sim$4k, which is negligible compared to the original 14.4M parameters.}
		\begin{tabular}{ccccccc}
			\toprule
			\multicolumn{5}{c}{Experimental setup} &
			\multicolumn{2}{c}{Downstream tasks} \\
			\cmidrule(lr){1-5}
			\cmidrule(lr){6-7}
			 Pace Pred. & Ctr. Learn. & Network & Configuration  &  Params &  UCF101 & HMDB51  \\
			\toprule
			\checkmark & $\times$ &R(2+1)D & - & 14.4M & \textbf{73.9} & \textbf{33.8} \\
			$\times$ & \checkmark &R(2+1)D & Same  pace & 14.4M & 59.4 & 20.3 \\
			$\times$ &  \checkmark &R(2+1)D & Same  context & 14.4M & 67.3 & 28.6 \\
			\midrule
			\checkmark & \checkmark &R(2+1)D & Same  pace & 14.4M & 73.6 & 32.3 \\
			\checkmark & \checkmark &R(2+1)D & Same  context & 14.4M & 75.8 & 35.0\\
			\checkmark & \checkmark & R(2+1)D + fc & Same context & 14.4M$^*$ & \textbf{75.9} & \textbf{35.9} \\
			\bottomrule
		\end{tabular}
		\label{tab:cl}
\end{table}

\subsubsection{Contrastive Learning.} The performances of the two contrastive learning configurations are shown in Table \ref{tab:cl}. Some key observations are listed for a better understanding of the contrastive learning: (1) The same pace configuration achieves much worse results than the same context configuration. We suspect the reason is that in the same pace configuration, as there are only four pace candidates $p=[1,4]$,  video clips are tend to belong to the same pace. Therefore, compared with the same context configuration, much fewer negative samples are presented in the training batches, withholding the effectiveness of the contrastive learning. (2) Pace prediction task achieves much better performance compared to each of the two contrastive learning settings. This demonstrates the superiority of the proposed pace prediction task.

When combining the pace prediction task with contrastive learning, similar to the observation described above, regarding the same pace configuration, performance is slightly deteriorated and regarding the same context configuration, performance is further improved both on UCF101 and HMDB51 datasets.
It shows that appropriate multi-task self-supervised learning can further boost the performances, in consistent with the observation in~\cite{doersch2017multi}.
Based on the same video content configuration, we further introduce a nonlinear layer between the embedding space and the final contrastive learning space to alleviate the direct influence on the pace prediction learning. It is shown that such a practice can further improve the performance (last row in Table~\ref{tab:cl}). 

\subsection{Action Recognition}

\begin{table}[tbp!]
\centering
	\caption{Comparison with the state-of-the-art self-supervised learning methods on UCF101 and HMDB51 dataset (Pre-trained on video modality only).$^*$The input video clips contain 64 frames.}
	\setlength{\tabcolsep}{2pt}
		\begin{tabular}{lcccccc}
			\toprule
			\multicolumn{1}{l}{\multirow{2}{*}{Method}} &
			\multicolumn{4}{c}{Pre-training settings} &
			\multicolumn{2}{c}{Evaluation} \\
			\cmidrule(lr){2-5}
			\cmidrule(lr){6-7}
			& Network  & Input size &  Params & Dataset&   UCF101 & HMDB51  \\
			\midrule
			Fully supervised  & S3D-G & $224\times224^*$ & 9.6M & ImageNet & 86.6 & 57.7 \\
			Fully supervised  & S3D-G & $224\times224^*$ &  9.6M & K-400 & 96.8 & 74.5 \\
			\midrule
			Object Patch\cite{wang2015unsupervised} & AlexNet  & $ 227 \times 227$ & 62.4M & UCF101 & 42.7 & 15.6\\
			ClipOrder\cite{misra2016shuffle} & CaffeNet  & $ 227 \times 227$ & 58.3M & UCF101 & 50.9 & 19.8\\
			Deep RL\cite{buchler2018improving} & CaffeNet  & $227 \times 227$ & - & UCF101  & 58.6 & 25.0\\
			OPN \cite{lee2017unsupervised}  & VGG &  $80 \times 80$ & 8.6M & UCF101 &59.8 & 23.8 \\
			VCP \cite{luo2020video} & R(2+1)D & $112\times112$ & 14.4M & UCF101 &  66.3 & 32.2\\ 
			VCOP \cite{xu2019self} & R(2+1)D & $112\times112$ & 14.4M & UCF101 &  72.4 & 30.9 \\
			PRP \cite{yao2020video} & R(2+1)D & $112\times112$ & 14.4M & UCF101 &  72.1 & 35.0 \\
			\textbf{Ours} & R(2+1)D  & $112\times112$ & 14.4M & UCF101 & \textbf{75.9} & \textbf{35.9}\\
			\midrule 
			MAS\cite{wang2019self} & C3D & $112\times112$ & 27.7M & K-400 & 61.2 & 33.4 \\
			RotNet3D \cite{jing2018self} & R3D-18 & $224\times224$& 33.6M & K-400 &  62.9 & 33.7 \\ 
			ST-puzzle \cite{kim2018self} & R3D-18 & $224\times224$ & 33.6M & K-400 &  65.8 & 33.7 \\
			DPC \cite{han2019video} & R3D-18 & $128\times128$ & 14.2M & K-400 &  68.2 & 34.5 \\
			DPC \cite{han2019video} & R3D-34 & $224\times224$ & 32.6M & K-400 &  75.7 & 35.7 \\
			\textbf{Ours} & R(2+1)D  & $112\times112$ & 14.4M & K-400 & \textbf{77.1} & \textbf{36.6} \\
			\midrule
			SpeedNet~\cite{benaim2020speednet} & S3D-G & $224\times224^*$ & 9.6M  & K-400 & 81.1 & 48.8 \\
			\textbf{Ours} & S3D-G & $224\times224^*$ & 9.6M & UCF101 & \textbf{87.1} & \textbf{52.6} \\
			\bottomrule
		\end{tabular}
	\label{tab:sota}
\end{table}

We compare our approach with other methods on the action recognition task in Table \ref{tab:sota}.We have the following key observations: (1) Our method achieve the state-of-the-art results on both UCF101 and HMDB51 dataset. When pre-trained on UCF101, we outperform the current best-performing method PRP~\cite{yao2020video}. When pre-trained on K-400, we outperform the current best-performing method DPC~\cite{han2019video}. 
(2) Note that here the DPC method uses R3D-34 as their backbone network and the video input size is $224 \times 224$ while we only use $112 \times 112$. When the input size of DPC is at the same scale as ours, \ie, $128\times128$, we outperform it by 8.9\% on UCF101 dataset.  We attribute such success to both our pace prediction task and the usage of R(2+1)D. It can be observed that with R(2+1)D and only UCF101 as pre-train dataset, VCOP~\cite{xu2019self} can achieve 72.4\% on UCF101 and 30.9\% on HMDB51. 
(3) Backbone networks, input size and clip length do play important roles in the self-supervised video representation learning. As shown in the last row, by using the S3D-G~\cite{xie2018rethinking} architecture with 64-frame clips as inputs, pre-training only on UCF101 can already achieve remarkable performance, even superior to fully supervised pre-training on ImageNet (on UCF101).

\begin{figure}[t]
    \centering
	\includegraphics[width=\textwidth]{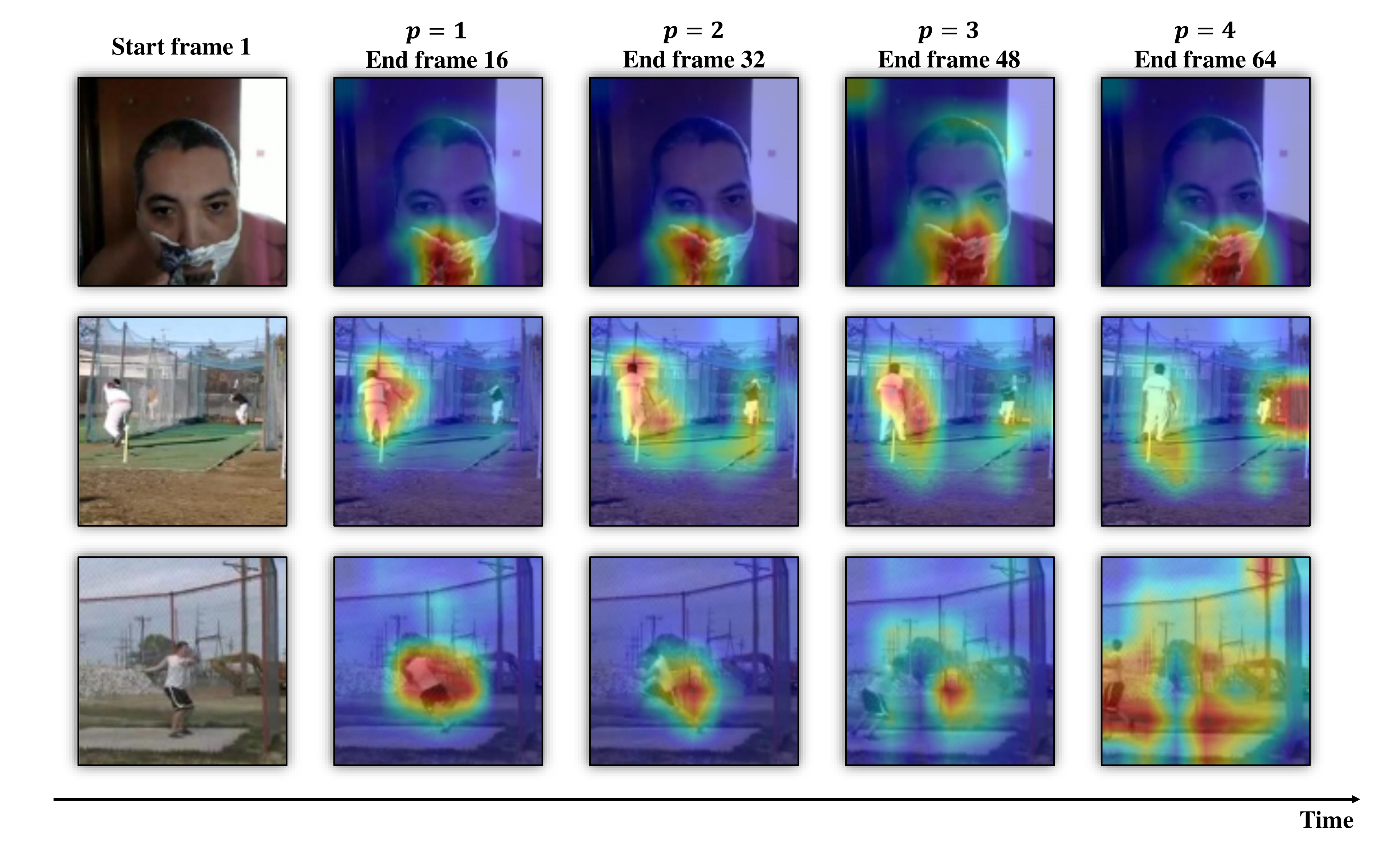}
	\caption{Attention visualization of the conv5 layer from self-supervised pre-trained model using \cite{zagoruyko2016paying}. Attention map is generated with 16-frames clip inputs and applied to the last frame in the video clips. Each row represents a video sample while each column illustrates the end frame \emph{w.r.t.} different sampling pace $p$.}
	\label{fig:att}
\end{figure}

To further validate the proposed approach, we visualize the attention maps based on the pre-trained R(2+1)D model, as shown in Fig. \ref{fig:att}. It can be seen from the attention maps that the neural network will pay more attention to the motion areas when learning the pace prediction task. It is also interesting to note that in the last row, as attention map on $p=4$ computes the layer information spanning 64 frames, it is activated at several motion locations.

\subsection{Video Retrieval}

\begin{table}[t]
	\begin{minipage}[t]{.49\textwidth}
	\caption{Comparison with state-of-the-art methods for nearest neighbour retrieval task on UCF101 dataset.}
	\label{tab:ucf101_fea}
    \centering
        \begin{adjustbox}{max width=\textwidth}
		\begin{tabular}{c|lccccc}
			\toprule
			~&Method  & Top1 & Top5 & Top10 & Top20  & Top50  \\
			\midrule
			\multirow{3}*{\rotatebox{90}{AlexNet}}&Jigsaw\cite{noroozi2016unsupervised} & 19.7 & 28.5 & 33.5 & 40.0 & 49.4 \\
			~&OPN\cite{lee2017unsupervised} & 19.9  & 28.7 & 34.0 & 40.6 & 51.6\\
			~&Deep RL\cite{buchler2018improving} & \underline{25.7} & 36.2 & 42.2 & 49.2 & 59.5 \\
			\midrule
			\multirow{5}*{\rotatebox{90}{C3D}}&Random& 16.7 & 27.5 & 33.7 & 41.4 & 53.0  \\
			~ & VCOP\cite{xu2019self} & 12.5 & 29.0 & 39.0 & 50.6 & 66.9\\
			~ & VCP\cite{luo2020video} & 17.3 & 31.5 & 42.0 & 52.6 & 67.7 \\
			~ & Ours (p5) & 20.0 & 37.4 & 46.9 & 58.5 & 73.1 \\
			~ & \textbf{Ours(p4)} & \textbf{31.9} & \textbf{49.7} & \textbf{59.2} & \textbf{68.9} & \textbf{80.2} \\
			\midrule
			\multirow{5}*{\rotatebox{90}{R3D-18}}&Random & 9.9 & 18.9 & 26.0 & 35.5 & 51.9  \\
			~&VCOP\cite{xu2019self} & 14.1 & 30.3 & 40.4 & 51.1 & 66.5\\
			~&VCP\cite{luo2020video} & 18.6 & 33.6 & 42.5 & 53.5 & 68.1 \\
			~ & Ours (p5) & 19.9 & 36.2 & 46.1 & 55.6 & 69.2 \\
			~&\textbf{Ours(p4)} & \textbf{23.8} & \textbf{38.1} & \textbf{46.4} & \textbf{56.6} & \textbf{69.8} \\
			\midrule
			\multirow{5}*{\rotatebox{90}{R(2+1)D}}& Random & 10.6 & 20.7 & 27.4 & 37.4 & 53.1  \\
			~ & VCOP\cite{xu2019self} & 10.7 & 25.9 & 35.4 & 47.3 & 63.9\\
			~ & VCP\cite{luo2020video} & 19.9 & 33.7 & 42.0 & 50.5 & 64.4 \\
			~ & Ours (p5) & 17.9 & 34.3 & 44.6 & 55.5 & 72.0 \\
			~&\textbf{Ours(p4)} & \textbf{25.6} & \textbf{42.7} & \textbf{51.3} & \textbf{61.3} & \textbf{74.0} \\
			\bottomrule
		\end{tabular}%}
		\end{adjustbox}
\end{minipage}
\hfill
	\begin{minipage}[t]{.49\textwidth}
	\caption{Comparison with state-of-the-art methods for nearest neighbor retrieval task on HMDB51 dataset.}
	\label{tab:hmdb51_fea}
    \centering
        \begin{adjustbox}{max width=\textwidth}
		\begin{tabular}{c|lccccc}
			\toprule
			&Method& Top1 & Top5 & Top10 & Top20  & Top50  \\
			\midrule
			\multirow{4}*{\rotatebox{90}{C3D}}&Random & 7.4 & 20.5 & 31.9 & 44.5 & 66.3  \\
			~& VCOP\cite{xu2019self}& 7.4 & 22.6 & 34.4 & 48.5 & 70.1\\
			~& VCP\cite{luo2020video}& 7.8 & 23.8 & 35.5 & 49.3 & 71.6 \\
			~ & Ours (p5) & 8.0 & 25.2 & 37.8 & 54.4 & 77.5 \\
			~ & \textbf{Ours(p4)} & \textbf{12.5} & \textbf{32.2} & \textbf{45.4} & \textbf{61.0} & \textbf{80.7} \\
			\midrule
			\multirow{5}*{\rotatebox{90}{R3D-18}}&Random & 6.7 & 18.3 & 28.3 & 43.1 & 67.9  \\
			~&VCOP\cite{xu2019self} & 7.6 & 22.9 & 34.4 & 48.8 & 68.9\\
			~&VCP\cite{luo2020video} & 7.6 & 24.4 & 36.6 & 53.6 & 76.4 \\
			~ & Ours (p5) & 8.2 & 24.2 & 37.3 & 53.3 & 74.5 \\
			~ & \textbf{Ours(p4)} & \textbf{9.6} & \textbf{26.9} & \textbf{41.1} & \textbf{56.1} & \textbf{76.5} \\
			\midrule
			\multirow{5}*{\rotatebox{90}{R(2+1)D}}& Random & 4.5 & 14.8 & 23.4 & 38.9 & 63.0  \\
			~&VCOP\cite{xu2019self} & 5.7 & 19.5 & 30.7 & 45.8 & 67.0\\
			~&VCP\cite{luo2020video} & 6.7 &  21.3 & 32.7 & 49.2 & 73.3\\
			~ & Ours (p5) & 10.1 & 24.6 & 37.6 & 54.4 & 77.1 \\
			~ & \textbf{Ours(p4)} & \textbf{12.9} & \textbf{31.6} & \textbf{43.2} & \textbf{58.0} & \textbf{77.1} \\
			\bottomrule
		\end{tabular}
		\end{adjustbox}
\end{minipage}
\end{table}

We further validate the proposed approach on the video retrieval task. Basically, we follow the same evaluation protocol described in \cite{luo2020video,xu2019self}. Ten 16-frames clips are sampled from each video and then go through a feed-forward pass to generate features from the last pooling layer (p5). 
For each clip in the testing split, the Top\emph{k} nearest neighbors are queried from the training split by computing the cosine distances between every two feature vectors.
%If the test clip class label is within the Top\emph{k} retrieval results, it is considered as correct. And the final accuracy is computed by averaging all the test results using the pooling layer with the best performance. 
We consider \emph{k} to be 1, 5, 10, 20, 50. To align the experimental results with prior works~\cite{luo2020video,xu2019self} for fair comparison, we use pre-trained models from the pace prediction task on UCF101 dataset.
As shown in Table~\ref{tab:ucf101_fea} and Table~\ref{tab:hmdb51_fea}, our method outperforms the VCOP~\cite{xu2019self} and VCP~\cite{luo2020video} in most cases on the two datasets across the three backbone networks. In practice, we also find that significant improvement can be achieved by using the second last pooling layer (p4).

\section{Conclusion}

In this paper, we proposed a new perspective towards self-supervised video representation learning, by pace prediction. Contrastive learning was also incorporated to further encourage the networks to learn high-level semantic features. To validate our approach, we conducted extensive experiments across several network architectures on two different downstream tasks. The experimental results demonstrated the superiority of our method on learning powerful spatio-temporal representations. Besides, the pace prediction task does not rely on any motion channel as prior information/input during training. As a result, such a pace prediction task can serve as a simple yet effective supervisory signal when applying the self-supervised video representation learning in real world, taking advantage of billions of video data freely. \\

\noindent\textbf{Acknowledgements.}
This work was partially supported by the HK RGC TRS under T42-409/18-R, the VC Fund 4930745 of the CUHK T Stone Robotics Institute, CUHK, and the EPSRC Programme Grant Seebibyte EP/M013774/1.

\par\vfill\par
% Now we have reached the maximum size of the ECCV 2020 submission (excluding references).
% References should start immediately after the main text, but can continue on p.15 if needed.

\bibliographystyle{splncs04}
\bibliography{egbib}

	% \linenumbers
	%\pagestyle{headings}
	%\mainmatter
	%\def\ECCVSubNumber{2819}  % Insert your submission number here
	
	\title{{Supplementary Materials 
	}} % Replace with your title
	
	\titlerunning{Supplementary Materials}
	\author{ }
	\institute{}
%	\author{Jiangliu Wang\inst{1} \and
%		Jianbo Jiao\inst{2} 
%		\and
%		{Yun-Hui Liu\inst{1}}}
%	%
%	\authorrunning{J. Wang et al.}
%	% First names are abbreviated in the running head.
%	% If there are more than two authors, 'et al.' is used.
%	
%	\institute{The Chinese University of Hong Kong
%		\and
%		University of Oxford\\
%		\email{\{jlwang,yhliu\}@mae.cuhk.edu.hk~}, \email{jianbo@robots.ox.ac.uk}}
	%******************
	\maketitle
	
	\section{Overview}
	
	In this supplementary material we provide:
	\begin{itemize}
		\item More ablation studies on the pace prediction task design in Sec. \ref{sec:ablation}
		\item Algorithm implementation details in Sec.~\ref{sec:alg}
		\item More qualitative results of attention maps with different paces in Sec. \ref{sec:vis}
		%\item A video file \textbf{suppleVideo.mp4} illustrating the basic idea of pace prediction. Note that we randomly sample videos with different paces from the UCF101 dataset~\cite{soomro2012ucf101}.
		%See Table \ref{tab:pace} for more details on the pace prediction accuracy. 
	\end{itemize}
	
	\section{Additional Ablation Studies on Pace Prediction Task} \label{sec:ablation}
	
	Here we provide additional ablation studies on the design of the pace prediction task, including (1) Pace prediction performance (Table~\ref{tab:pace}). (2) Evaluation of the performance on \emph{slow} pace as described in our paper (Table~\ref{tab:slow}). (3) Investigation on different pace steps (Table~\ref{tab:step}). (4) Analysis on video play direction, \ie, forwards or backwards (Table~\ref{tab:back}). 
	
	\textbf{Pace prediction accuracy.} We report the pretext task performance (\ie, pace prediction accuracy) and the downstream task performance (\ie, action recognition accuracy) on UCF101 dataset in Table \ref{tab:pace}. It can be seen from the table that with the increase of the maximum pace, the pretext task becomes harder for the network to solve, which leads to degradation of the downstream task. This further validate our claim in the paper that a pretext task should be neither too simple nor too ambiguous.   
	
	\begin{table}
		\centering
		\caption{Pace prediction accuracy w.r.t. different pace design. }
		\label{tab:pace}
		\centering
		\begin{tabular}{ccccc}
			\toprule
			Pre-training & Method & \# Classes & Pace rea. acc. &UCF acc.\\
			\midrule
			$\times$  & Random & - & - &56.0 \\
			\midrule
			% 			\cmidrule{2-4}
			\checkmark  & $p=[1,3]$ & 3 &  77.6 &71.4 \\
			\checkmark  & $p=[1,4]$ &  4 &69.5 &\textbf{72.0} \\
			\checkmark  & $p=[1,5]$ & 5 &61.4 &72.0 \\
			\checkmark  & $p=[1,6]$ & 6 &55.9 & 71.1 \\
			
			%\midrule
			%\checkmark & + color jittering & 53.9 & \textbf{73.9} \\
			\bottomrule
		\end{tabular}
		\label{tab:pace}
	\end{table}

	\textbf{Slow pace.} In our paper, we propose two different methods to generate video clips with slow pace: replication of previous frames or interpolation with existing algorithms~\cite{ref1}. We choose the replication in practice as most modern interpolation algorithms are based on supervised learning, while our work focuses on self-supervised learning, forbidding us to use any human annotations.
	
	As shown in Table \ref{tab:slow}, compared with normal and \emph{fast} paces,  if we use normal and \emph{slow} paces, the performance of the downstream task decreases (73.9$\rightarrow$72.6). While when combining with both slow and fast pace (\emph{absolute} pace as described in the paper), no performance change is observed, which again validates our choice of the pace configuration.
	
	\begin{table}
		\centering
		%\caption{Comparison with the state-of-the-art self-supervised video representation learning methods on UCF101 and HMDB51 dataset.}
		\caption{Evaluation of slow pace.}\vspace{-3mm}
		% 	\begin{center}
		\setlength{\tabcolsep}{5pt}
		\begin{tabular}{cccc}
			\toprule
			Config. & Pace & \# Classes & UCF10 Acc.\\
			\toprule
			Baseline & [1,2,3,4]  & 4 & 73.9\\
			\midrule
			Slow & [$\frac{1}{4}$,$\frac{1}{3}$,$\frac{1}{2}$,1]  & 4 & 72.6\\
			\midrule
			Slow-fast & [$\frac{1}{3}$,$\frac{1}{2}$,1,2,3]  & 5 & 73.9 \\
			\bottomrule
		\end{tabular}
		\label{tab:cl}%\vspace{-3mm}
		% 	\end{center}
		\label{tab:slow}
	\end{table}

	\textbf{Pace step.} Based on the better performance achieved by the fast pace as shown above, we take a closer look into the fast pace design, by considering different interval steps, \ie, frame skip. For simplicity, in the paper we showcase with the step that equals one (baseline) between each fast pace where the paces are \{1,2,3,4\}. Here we further explore the interval steps of two and three so as to introduce larger motion dynamics into the learning process. It can be observed from Table~\ref{tab:step} that by increasing the interval steps, performance could be further improved, but tends to  saturate when the step is too large.         
	\begin{table}
		\centering
		%\caption{Comparison with the state-of-the-art self-supervised video representation learning methods on UCF101 and HMDB51 dataset.}
		\caption{Evaluation of different pace steps.}\vspace{-3mm}
		% 	\begin{center}
		\setlength{\tabcolsep}{5pt}
		\begin{tabular}{cccc}
			\toprule
			Step & Pace & \# Classes & UCF10 Acc.\\
			\toprule
			1 & [1,2,3,4]  & 4& 73.9\\
			\midrule
			2 & [1,3,5,7]  & 4 & 74.9\\
			\midrule
			3 & [1,4,7,10] & 4 & 74.7 \\
			\bottomrule
		\end{tabular}
		\label{tab:cl}%\vspace{-3mm}
		% 	\end{center}
		\label{tab:step}
	\end{table}

	\textbf{Forwards \emph{v.s.} backwards.} It has been a long standing problem that whether a forward played video can be considered as the same as its backward played version, in self-supervised video representation learning. Some works~\cite{ref2,ref3} argue that these two versions should be attributed to the same semantic labels, while Wei \etal~prone to distinguish the forwards and backwards video~\cite{ref5}. In the following, we investigate these two opinions based on our method as shown in Table \ref{tab:back}.

	As for the random backwards with four classes, we consider forwards and backwards videos as the same pace samples, while for backwards with eight classes, they are considered to be different samples. It can be seen from the table that, both configurations achieve lower performance than our baseline. We suspect the reason is that to distinguish the backwards from forwards, it is essentially a video order prediction task though in some order prediction work~\cite{ref2,ref3} they are considered to be the same. When combing the proposed pace prediction task with such an order prediction task, the network will be confused towards an ambiguous target. As a result, the downstream task performance is deteriorated.     
	
	\begin{table}
		\centering
		%\caption{Comparison with the state-of-the-art self-supervised video representation learning methods on UCF101 and HMDB51 dataset.}
		\caption{Evaluation of video forwards \emph{v.s.} backwards.}\vspace{-3mm}
		% 	\begin{center}
		\setlength{\tabcolsep}{5pt}
		\begin{tabular}{cccc}
			\toprule
			Config. & Pace & \# Classes&  UCF10 Acc.\\
			\toprule
			Baseline  &[1,2,3,4] & 4 & 73.9\\
			\midrule
			Rnd backwards & [1,2,3,4] & 4 & 73.0\\
			\midrule
			Backwards & [$\pm1$, $\pm2$, $\pm3$, $\pm4$] & 8 &  73.7 \\
			%\midrule
			%Backwards in the same batch & [$\pm1$, $\pm2$, $\pm3$, $\pm4$] & 8 & 38.6 & 76.5 \\
			\bottomrule
		\end{tabular}
		\label{tab:cl}%\vspace{-3mm}
		% 	\end{center}
		\label{tab:back}
	\end{table}

	\section{Implementation Details} \label{sec:alg}
	
	Here we present the algorithms of the proposed approach, with two possible solutions as mentioned in the paper: pace prediction with contrastive learning on same video context and pace prediction with contrastive learning on same video pace.

	\begin{algorithm}[h]
		\caption{Pace prediction with contrastive learning on same video context.}
		\hspace*{0.02in} {\bf Input:} Video set X, pace transformation $g_{pac}(.)$, $\lambda_{cls}$, $\lambda_{ctr}$, backbone network $f$.
		\\
		\hspace*{0.02in} {\bf Output:} Updated parameters of network $f$.
		\begin{algorithmic}[1]
			\For{sampled mini-batch video clips \{$x_1, \dots, x_N$\} }
			\For{$i=1$ to $N$}
			\State Random generate video pace $p_i$, ${p_i}'$
			\State $\widetilde{x}_i=g_{pac}(x_i|p_i)$
			\State ${\widetilde{x}_i}'=g_{pac}(x_i|{p_i}')$
			\State $z_i = f(\widetilde{x}_i)$
			\State ${z_i}' = f({\widetilde{x}_i}')$
			\EndFor
			\For {$i\in \{1, \dots, 2N\}$ and $j\in \{1, \dots, 2N\}$}
			\State$\mysim(z_i, z_j)={z_i}^\top z_j$		
			\EndFor
			
			\State Define $\mathcal{L}_{ctr\_sc} = -\frac{1}{2N} \sum\limits_{i,\mathcal{J}} \log \frac{\exp (\mysim(z_i, {z_i}'))}{\sum\limits_{i} \exp(\mysim(z_i, {z_i}'))+ \sum\limits_{i,\mathcal{J}} \exp(\mysim(z_i, z_\mathcal{J}))}.$ \\
			\State$\mathcal{L}_{cls} = -\frac{1}{2N}\sum\sum\limits_{i=1}^{M}y_i(\log\frac{\exp(h_i)}{\sum_{j=1}^{M}\exp(h_j)})$
			
			\State $\mathcal{L} = \lambda_{cls}\mathcal{L}_{cls} + \lambda_{ctr}\mathcal{L}_{ctr\_sc}$
			
			\State Update $f$ to minimize $\mathcal{L}$
			\EndFor
		\end{algorithmic}	
	\end{algorithm}
	
	\begin{algorithm}[h]
		\caption{Pace prediction with contrastive learning on same video pace. }
		\hspace*{0.02in} {\bf Input:} Video set X, pace transformation $g_{pac}(.)$, $\lambda_{cls}$, $\lambda_{ctr}$, backbone network $f$.
		\\
		\hspace*{0.02in} {\bf Output:} Updated parameters of network $f$.
		\begin{algorithmic}[1]
			\For{sampled mini-batch video clips \{$x_1, \dots, x_N$\} }
			\For{$i=1$ to $N$}
			\State Random generate video pace $p_i$
			\State $\widetilde{x}_i=g_{pac}(x_i|p_i)$
			\State $z_i = f(\widetilde{x}_i)$
			\EndFor
			\For {$i\in \{1, \dots, N\}$ and $j\in \{1, \dots, N\}$}
			\State$\mysim(z_i, z_j)={z_i}^\top z_j$		
			\EndFor
			
			\State Define $\mathcal{L}_{ctr\_sp} = -\frac{1}{N} \sum\limits_{i,j,k} \log \frac{\exp (\mysim(z_i, z_j))}{\sum\limits_{i,j} \exp(\mysim(z_i, z_j))+ \sum\limits_{i,k} \exp(\mysim(z_i, z_k))}$ \\
			\State$\mathcal{L}_{cls} = -\frac{1}{N}\sum\sum\limits_{i=1}^{M}y_i(\log\frac{\exp(h_i)}{\sum_{j=1}^{M}\exp(h_j)})$
			
			\State $\mathcal{L} = \lambda_{cls}\mathcal{L}_{cls} + \lambda_{ctr}\mathcal{L}_{ctr\_sp}$
			
			\State Update $f$ to minimize $\mathcal{L}$
			
			\EndFor
		\end{algorithmic}	
	\end{algorithm}

	\section{Attention Visualization}
	\label{sec:vis}
	
	Finally, we provide the attention map visualization on more video clips with different paces. Starting from the same initial frame, we sample 16-frames clips with different paces $p=1,2,3,4$. Then we show the attention maps for every 3 frames. Note that only one attention map is generated based on a 16-frame video clip. It can be seen from Fig. \ref{fig:att}, clips with larger pace $p$ contain larger motion dynamics as they span more frames. The attention maps are also becoming active in larger motion areas with the increase of pace $p$.

	\begin{figure}[t]
		\centering
		\includegraphics[width=\textwidth]{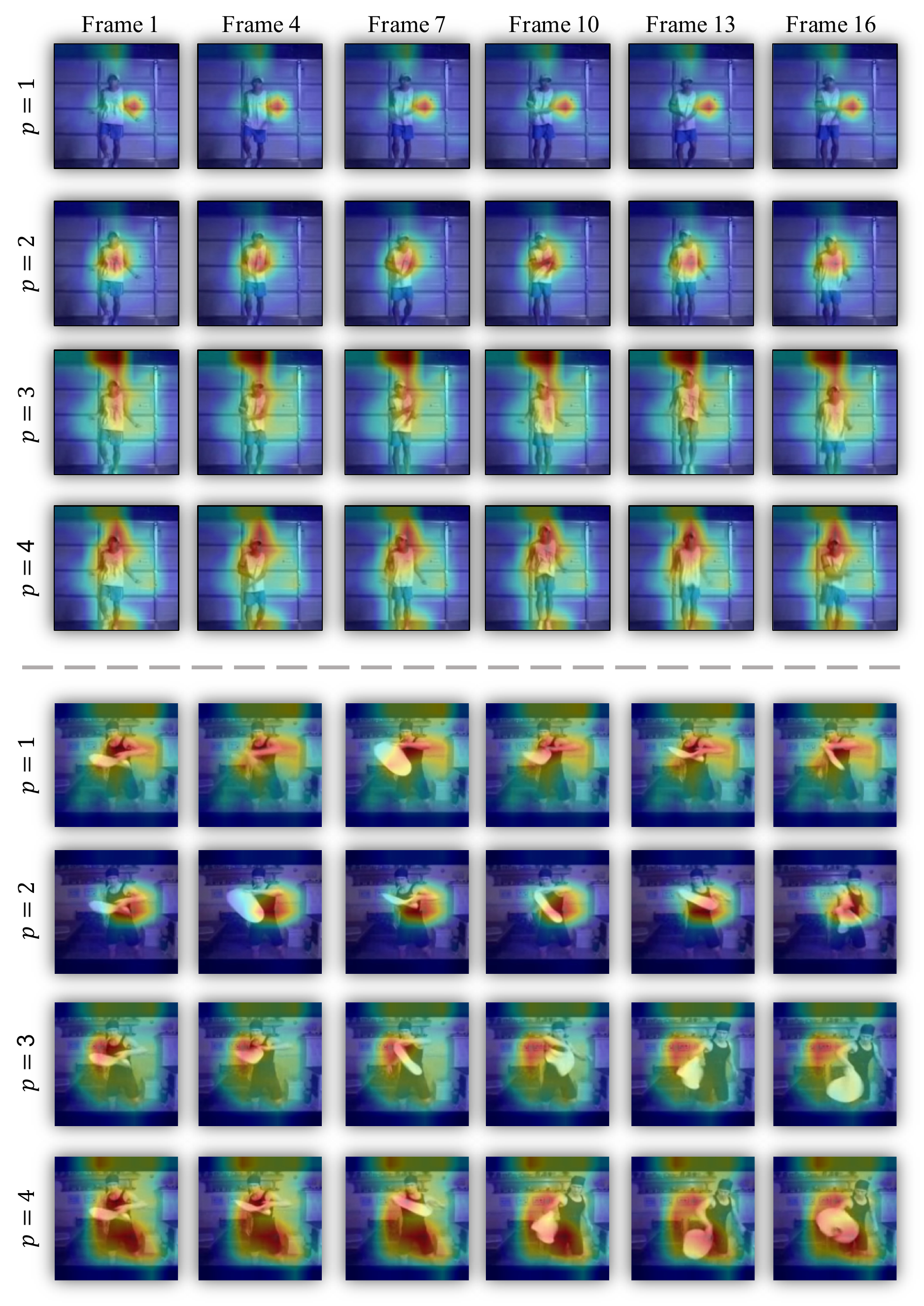}\vspace{-3mm}
		\caption{Attention visualization (using tool from \cite{ref6}) of the conv5 layer from self-supervised pre-trained model. }
		% 	\vspace{-3mm}
		\label{fig:att}
	\end{figure}
	

\end{document}